# Executable Epistemology: The Structured Cognitive Loop as an Architecture of Intentional Understanding

Myung Ho Kim
Professor, JEI University
enkiluv@gmail.com

## Abstract

Large language models exhibit intelligence without genuine epistemic understanding, revealing a fundamental philosophical gap: the absence of epistemic architecture. This paper introduces the Structured Cognitive Loop (SCL) as an executable epistemological framework for emergent intelligence.

Unlike traditional AI research that asks "what is intelligence?" (ontological), SCL asks "under what conditions does cognition emerge?" (epistemological). Situated within contemporary philosophy of mind and cognitive phenomenology, this framework bridges conceptual philosophy and implementable cognition. Drawing on process philosophy, enactive cognition, and extended mind theory, we reconceptualize intelligence not as a possessed property but as a performed process - a continuous loop of cognition, control, action, and memory structured by normative regulation.

SCL makes three interrelated contributions. First, it operationalizes philosophical insights into computationally interpretable structures, enabling what we term "executable epistemology" - philosophy as structural experiment. Second, it demonstrates that functional separation within cognitive architecture yields more coherent and interpretable behavior than monolithic prompt-based approaches, with empirical support from controlled agent evaluations. Third, it redefines the measure of intelligence: not representational accuracy but the capacity to reconstruct one's own epistemic state through intentional understanding.

This framework has implications across philosophy of mind, epistemology, and artificial intelligence. For philosophy of mind, it offers a new mode of engagement where theories of cognition can be enacted and tested. For AI, it grounds behavioral intelligence in epistemic structure rather than statistical regularity. For epistemology, it suggests that knowledge is best understood not as truth-possession but as continuous structural reconstruction within a phenomenologically coherent loop.

We situate SCL within debates on cognitive phenomenology, emergence, normativity, and intentionality, arguing that genuine progress requires not larger models but architectures that structurally realize cognitive science principles.

**Keywords:** philosophy of mind, cognitive phenomenology, epistemology, artificial intelligence, cognitive architecture, process philosophy, enactive cognition, executable philosophy, intentional understanding.

## 1. Introduction

Large language models exhibit fluent text generation, solve complex problems, and engage in sophisticated reasoning, yet they operate without what philosophers call epistemic understanding - a concept that, in analytic epistemology, aligns with reflective understanding as theorized in virtue and structural epistemology (Greco, 2010; Kelp, 2015; Carter & Pritchard, 2021). Specifically, this is the capacity to reconstruct the structural pathway from evidence to conclusion. This gap is not simply John Searle's (1980) syntax-semantics distinction revisited, but a different gap: the absence of epistemic architecture. While Searle argued that computation alone cannot yield intentionality, we identify a prior condition: systems require structural relations among cognition, action, and memory before intentional understanding becomes possible. Current AI systems lack not semantics but the architectural conditions that enable epistemic coherence. This paper argues that addressing this gap requires shifting from ontological questions about what intelligence is to epistemological questions about under what structural conditions cognition emerges. To address this, we introduce the Structured Cognitive Loop (SCL), an executable framework for exploring these epistemic conditions.

This gap, we argue, reflects a persistent ontological bias in artificial intelligence research. From symbolic AI's equation of intelligence with symbol manipulation (Newell & Simon, 1976), through connectionism's identification with pattern recognition (Rumelhart & McClelland, 1986), to contemporary deep learning's reliance on statistical approximation (LeCun et al., 2015), the field has consistently treated intelligence as a determinate property awaiting instantiation. The question has always been "what is intelligence and how do we build it?" rather than "under what conditions does intelligent behavior emerge?" This bias has philosophical roots in what we might call substance ontology - the view that mental phenomena are properties locatable within discrete entities (Gallagher & Zahavi, 2021). Cognitive science exhibits parallel tendencies: working memory theorized as a capacity (Baddeley, 2012), executive function as a system (Diamond, 2013), intelligence as a measurable property indexed by psychometric tests (Deary, 2020). While these approaches have yielded valuable insights, they share what Gilbert Ryle (1949) called a category mistake: treating dispositional concepts as if they named occurrent states or intrinsic properties. Intelligence, on this alternative view, is not something a system has but something it does - not a substance but a mode of organization, not a property but a performance.

An alternative framework emerges from convergent insights in process philosophy, enactive cognition, and extended mind theory. Contemporary process metaphysics reconceives entities as events characterized by their relations rather than substances characterized by their properties (Rescher, 2000; Seibt, 2018). Enactive approaches argue that cognition arises through structural coupling between organism and environment, not through internal representation (Thompson, 2007; Di Paolo et al., 2017). Extended mind theory demonstrates that cognitive processes routinely incorporate external tools and scaffolds, such that the boundaries of mind extend beyond biological brains (Clark & Chalmers, 1998; Clark, 2008). Despite their differences, these traditions converge on a core insight: cognition is relational rather than substantial, processual rather than static, distributed rather than localized. Recent work in cognitive phenomenology has further articulated this view, arguing that understanding has a distinctive experiential character that cannot be reduced to perceptual or affective qualities but emerges from the structural coherence of cognitive states (Kriegel, 2015; Smithies, 2019).

This convergence motivates what we call an epistemological turn. Following Kant's critical philosophy, which asked not "what is the world?" but "under what conditions is experience of the world possible?" (Kant, 1781/1998), we ask not "what is intelligence?" but "under what structural conditions does epistemic understanding emerge?" This parallels contemporary discussions in philosophy of science concerning the structural conditions of explanation (Woodward, 2003; Strevens, 2008), situating cognition as a case of epistemic structuring rather than mere representational mapping. This shift reframes the problem: instead of identifying where intelligence resides, we articulate the relational architecture that enables a system to maintain coherence between perception, memory, cognition, and action - what phenomenologists call intentional understanding (Zahavi, 2005; Gallagher, 2017). Such architecture is not a container but a set of structural relations; not a mechanism but an organization; not something a system possesses but something it enacts through continuous loops of self-correction and reconstruction. This perspective resonates with recent proposals for "architectural" approaches to artificial intelligence that emphasize modularity, transparency, and principled design over scale alone (Goertzel, 2014; Laird et al., 2017; Lake et al., 2017).

The Structured Cognitive Loop (SCL) operationalizes these philosophical insights. It reconceives intelligence not as a property of isolated components but as an emergent pattern arising from five interacting functions: cognition (inference grounded in evidence), control (enforcement of epistemic norms), action (environmental coupling through tools), memory (persistent structured state) governed by regulation (explicit epistemic norms and reasoning rules). These are not modules in a traditional computational sense but relational nodes whose continuous interaction produces what we recognize as understanding (Kim, 2025). The system "knows" not when it outputs correct answers but when it can reconstruct its own epistemic trajectory - citing evidence, explaining decisions, and revising judgments based on feedback. A companion technical study demonstrated that architectural separation of these functions produces measurable improvements in coherence, interpretability, and goal fidelity compared to monolithic prompt-based approaches across controlled multi-step tasks (Kim, 2025). However, the present paper asks a deeper question: what makes this architecture epistemologically privileged? Why does separating cognition, control, action, and memory enable more reliable understanding? Our answer is that SCL embodies structural conditions for intentional understanding - conditions that cannot be achieved through scaling alone but require architectural commitments reflecting principles from cognitive science and philosophy of mind.

This paper makes three interrelated contributions. First, it operationalizes philosophical insights about cognition into computationally interpretable structures, demonstrating what we term executable epistemology - philosophy not as abstract theory but as structural experiment. This approach extends the scope of philosophical inquiry: theories of mind can be implemented, tested, and refined through observation of their enacted behavior (Froese & Ziemke, 2009; Froese et al., 2013). Methodologically, this situates SCL within what has been called philosophical engineering (Dennett, 1995; Piccinini, 2015), where philosophical hypotheses are explored through executable models rather than solely through argumentation. The idea is not to reduce philosophical concepts to code, but to test whether philosophical claims about cognitive structure remain coherent when made explicit and operational. Second, by grounding SCL in both empirical evaluation and conceptual analysis, we show that architectural modularity reflecting cognitive science principles produces not just improved performance metrics but qualitatively different forms of cognition - systems capable of epistemic self-awareness and structural self-explanation. Third, we redefine the measure of intelligence from representational accuracy to structural coherence: a system exhibits understanding when it maintains phenomenologically coherent relations among cognition, action, and memory over time. This shift has implications for debates on interpretability, where current approaches focus on post-hoc explanation rather than architectural transparency (Lipton, 2018; Rudin, 2019), and for philosophical discussions of intentionality, where the question is not whether artificial systems can have "original" intentionality but whether they can maintain the structural conditions that constitute understanding (Haugeland, 1998; Wheeler, 2005).

The paper proceeds as follows. Section 2 develops philosophical foundations, tracing the shift from classical ontologies of mind through process philosophy, enactive cognition, and cognitive phenomenology to the concept of epistemic architecture. Section 3 presents SCL's structure in detail, explaining how each function contributes to epistemic coherence and how their interaction produces emergent understanding. Section 4 explores implications for three key debates in philosophy of mind and AI: the nature of intentionality in artificial systems, the structural basis of normativity, and the relationship between architectural transparency and interpretability. Section 5 concludes with methodological reflections on executable epistemology as a mode of philosophical inquiry, considering both its promise and limitations.

The philosophical stakes are clear: if SCL succeeds as executable epistemology, it demonstrates that conceptual analysis and architectural design are complementary domains - philosophy engaging implementable structures without reducing to engineering.

## 2. Philosophical Background: From Ontology to Epistemology

The absence of epistemic architecture in current AI systems reflects deeper philosophical commitments about the nature of intelligence. This chapter traces a conceptual transition: from classical ontologies that treat intelligence as a possessed property to alternative frameworks - process philosophy, enactive cognition, extended mind, and cognitive phenomenology - that reconceive cognition as relational, processual, and distributed. These traditions converge on an epistemological turn: instead of asking "what is intelligence?" we ask "under what structural conditions does epistemic understanding emerge?" We conclude by identifying five such conditions that define epistemic architecture and provide the philosophical blueprint for SCL.

### 2.1 Classical Ontologies of Mind: From Rationalism to Computationalism

The ontological approach to intelligence has deep roots in modern philosophy. Descartes' substance dualism established a paradigm in which mind is conceived as a distinct substance characterized by intrinsic properties - thinking, willing, judging - locatable within a unified entity (Descartes, 1641/1996). The *res cogitans*, for Descartes, was defined by its capacity for thought, a property that existed independently of bodily processes. While subsequent materialist and functionalist revisions rejected Cartesian substance dualism, they retained the underlying ontological framework: mental phenomena as properties inhering in determinate substrates, whether neural tissues or computational states. This commitment persists in contemporary cognitive science and artificial intelligence, manifesting as the search for the "location" or "substrate" of intelligence - whether in synaptic weights, neural assemblies, or algorithmic structures.

The transition from rationalist metaphysics to computational models of mind preserved this ontological bias while shifting its locus. Early symbolic AI, exemplified by Newell and Simon's (1976) Physical Symbol System Hypothesis, treated intelligence as the capacity for symbol manipulation according to syntactic rules. Intelligence, on this view, resided in the system's ability to instantiate formal operations over discrete representations. The computer became a new substrate for mental properties, yet the underlying question remained ontological: *what kind of entity possesses intelligence?* Connectionism challenged symbolic AI's emphasis on discrete representations but retained the ontological framing (Rumelhart & McClelland, 1986). Neural networks were proposed as alternative substrates - distributed patterns of activation replacing localized symbols - yet the question remained one of location and instantiation. Intelligence was still conceived as a property awaiting realization in the right kind of material or formal structure.

Contemporary deep learning extends this trajectory. Large language models achieve remarkable linguistic competence through statistical learning over vast corpora (LeCun et al., 2015; Brown et al., 2020), yet the explanatory framework remains ontological: intelligence as emergent from scale, architecture, and training regime. The operative assumption is that sufficient parameters, data, and compute will yield intelligence as an intrinsic property of the resulting system. This perspective treats intelligence as something a model *has* - measurable through benchmarks, scalable through resources, locatable within learned representations. Even discussions of "emergence" in large models typically frame the phenomenon ontologically: new capacities appearing as system size crosses critical thresholds, as if intelligence were a phase transition in parameter space (Wei et al., 2023).

Cognitive science exhibits parallel tendencies. Working memory is theorized as a limited-capacity buffer (Baddeley, 2012), executive function as a control system (Diamond, 2013), attention as a selective filter (Posner & Petersen, 1990). Each framework posits mental faculties as determinate structures or resources - entities that can be measured, depleted, and enhanced. Intelligence itself becomes reified as $g$, the general factor extracted from psychometric tests, treated as a stable individual property predictive of cognitive performance (Deary, 2020). While these approaches have yielded valuable empirical insights, they share what Gilbert Ryle (1949) diagnosed as a category mistake: treating dispositional concepts - abilities, tendencies, capacities - as if they named occurrent states or intrinsic properties rather than patterns of behavior or organization.

The ontological bias creates characteristic explanatory gaps. If intelligence is a property, what explains its flexibility across contexts? If it resides in particular structures, why do different architectures produce functionally equivalent cognition? If it emerges from scale alone, why do systems fail catastrophically on seemingly simple tasks (Marcus, 2018)? These puzzles suggest that the ontological framework may be asking the wrong questions. Rather than seeking the substance or substrate of intelligence, we might instead ask: under what conditions does intelligent behavior emerge? What structural relations enable systems to maintain coherence between perception, memory, proposal, and action? This shift - from ontology to epistemology - reframes intelligence not as something possessed but as something enacted through organized process.

## 2.2 Process Philosophy and the Dynamic View of Cognition

Process philosophy offers a radical alternative to substance ontology. Rather than conceiving reality as composed of enduring entities with intrinsic properties, process thought reconceives being as becoming - a flux of events, relations, and transformations in which stability is an achievement rather than a given (Rescher, 2000; Seibt, 2018). For Alfred North Whitehead, the fundamental units of reality are not substances but "occasions of experience" - momentary events characterized entirely by their relations to other events (Whitehead, 1929/1978). An occasion has no intrinsic nature independent of its relations; it is what it does within a relational field. Properties, on this view, are not possessed but enacted through process.

Applied to cognition, process philosophy suggests that mind is not a container of mental states but an ongoing activity of self-organization. Whitehead's notion of "concrescence" - the process by which an occasion comes into being through the integration of its relations - provides a model for understanding how cognitive coherence emerges from the coordination of perception, memory, and action. Cognition is not the manipulation of representations by a pre-existing subject but the continuous production of subjectivity through relational activity. The mind does not have experiences; it is the process of experiencing, constituted moment-by-moment through its engagement with the world.

Henri Bergson's concept of *durée* - duration or lived time - further articulates this processual view (Bergson, 1889/2001). Against mechanistic models that treat time as a succession of discrete instants, Bergson argues that consciousness is characterized by continuity and interpenetration. Past and present are not separate moments but coexist in memory, which is not a storehouse of fixed representations but a living synthesis that continuously reconstructs the past in light of present concerns. Memory, for Bergson, is not re-presentation but re-enactment - a creative process rather than a retrieval operation. This challenges computational models that treat memory as information storage, suggesting instead that remembering is a form of becoming rather than accessing.

Contemporary process metaphysics extends these insights through formal frameworks. Johanna Seibt's (2018) "process ontology" provides categories for describing dynamic systems without reducing them to static entities. Processes are characterized by their topology (how parts relate), their dynamics (patterns of change), and their normativity (tendencies toward certain trajectories). Cognitive processes, on this view, exhibit distinctive topologies - recursive, self-referential, anticipatory - that cannot be adequately captured by entity-based ontologies. Understanding cognition requires describing how processes generate, maintain, and transform themselves over time.

These insights have profound implications for artificial intelligence. If cognition is fundamentally processual, then building intelligent systems requires not instantiating the right representations or architectures but creating the right kinds of *organizational dynamics*. Intelligence is not a property to be engineered into a system but an achievement that emerges from ongoing cycles of perception, action, and adaptation. This suggests that AI research should focus less on what intelligent systems are and more on what they do - the processes they enact, the relations they maintain, the transformations they undergo. A process-oriented approach treats intelligence as a verb rather than a noun, as performance rather than possession.

Process philosophy also illuminates the temporal structure of cognition. Whitehead's analysis of "prehension" - the way each occasion grasps and integrates its past - suggests that cognition is inherently retrospective and reconstructive. Current experience is not simply caused by prior states but actively synthesizes those states into a coherent present. This resonates with contemporary discussions of predictive processing, where perception involves the continuous integration of sensory input with memory-based predictions (Clark, 2016). The process view emphasizes that this integration is not computation over static representations but dynamic reconstruction - the past is not retrieved but re-created in the act of cognition.

Finally, process philosophy challenges the mind-world boundary. If entities are defined by their relations, then cognition cannot be isolated within individual brains or systems. The cognitive process extends into the environment, incorporating tools, social practices, and institutional structures as constitutive elements. This anticipates recent work on extended and distributed cognition, which we examine in the next section. The process view thus prepares the ground for understanding cognition as an architectural achievement - not the property of a substance but the organization of relations across boundaries traditionally thought to separate mind from world.

## 2.3 Enactive and Extended Cognition: Mind as Interaction

The enactive approach to cognition, developed by Francisco Varela, Evan Thompson, and Eleanor Rosch, radically repositions the relationship between organism and environment (Varela et al., 1991; Thompson, 2007). Against representationalist models that conceive cognition as internal processing of external information, enactivism argues that cognition is *constituted* by sensorimotor interaction. Perception is not passive reception but active exploration; action is not output following computation but the very means by which meaning emerges. The organism does not represent a pre-given world but *enacts* a world through its pattern of engagements. Cognition, on this view, is not in the head but in the dynamic coupling between organism and environment.

Central to enactivism is the concept of *autopoiesis* - self-production or self-maintenance (Maturana & Varela, 1980). Living systems are characterized by circular organization: they produce the very components that constitute them, maintaining their identity through continuous metabolic activity. Cognition, for Varela, is the sense-making activity through which autonomous systems maintain their organization in relation to environmental perturbations. A cognitive system is not one that processes information but one that generates and sustains a perspective - a normative framework for distinguishing significant from insignificant perturbations. Cognition is thus inherently normative: it involves mattering, valuing, and goal-directed activity.

Enactivism emphasizes *structural coupling* - the co-evolution of organism and environment such that each shapes the other over time (Di Paolo et al., 2017). An organism's perceptual and motor capacities are not designed to represent an objective world but to maintain viability within its niche. The environment, in turn, is not a fixed backdrop but a domain of significance constituted by the organism's needs and capacities. This reciprocal determination means that cognition cannot be understood by analyzing either organism or environment in isolation; it exists in the relation between them. Perception is not information extraction but participatory sense-making - a process that simultaneously discloses a world and constitutes a self.

The extended mind thesis, articulated by Andy Clark and David Chalmers (1998), complements enactivism by demonstrating that cognitive processes routinely incorporate external resources. When Otto uses a notebook to store information that his biological memory cannot retain, the notebook functions as part of his cognitive system - it plays the same functional role as biological memory (Clark & Chalmers, 1998). The boundaries of cognition, Clark argues, are not fixed by skin or skull but are determined functionally: whatever plays the right causal role in producing intelligent behavior counts as part of the cognitive system. This principle of *parity* - if an external resource performs the same function as an internal process, it is part of the cognitive system - challenges internalist assumptions pervasive in cognitive science and AI.

Clark's later work extends this insight through the concept of cognitive scaffolding (Clark, 2008). Humans routinely off-load cognitive work onto environmental structures: we use written lists to organize tasks, diagrams to reason about spatial relations, cultural practices to coordinate collective action. These scaffolds are not mere aids to cognition but constitutive components - they change what we can think, not merely how efficiently we think it. The extended mind is thus not a fixed system but a flexible assemblage, continuously reconfiguring itself by incorporating new tools and practices. This has profound implications for understanding both human intelligence and artificial systems: intelligence is not located in brains or processors but distributed across heterogeneous networks of biological, technological, and social resources.

Shaun Gallagher's (2017) enactivist interventions further develop these themes by emphasizing the role of action and embodiment in shaping cognition. Against "brainbound" theories that treat the body as a peripheral input-output device, Gallagher shows that bodily structure and sensorimotor capacities fundamentally shape what we can perceive and understand. The body is not a container for cognition but a condition of possibility - the organized sensorimotor repertoire through which meaning emerges. This perspective challenges AI systems that treat embodiment as optional: if cognition is constituted through interaction, then disembodied systems cannot achieve the same form of understanding as embodied agents, regardless of their computational power.

The implications for artificial intelligence are significant. If cognition is enacted through structural coupling rather than representation, then building intelligent systems requires creating appropriate organism-environment relations rather than encoding world models. Intelligence emerges from interaction, not from internal complexity alone. This suggests that AI research should focus on designing systems capable of meaningful environmental engagement - systems that can maintain themselves through adaptive coupling rather than simply processing inputs. The enactive view also implies that cognition is fundamentally perspectival: there is no view from nowhere, no objective representation. Every cognitive system enacts a world from its own situated perspective, structured by its needs, capacities, and history.

Extended cognition further suggests that intelligence in AI need not be concentrated in a single model or architecture. Just as human cognition distributes itself across biological memory, notebooks, smartphones, and social practices, artificial intelligence might be reconceived as distributed assemblages in which different components - language models, external memory systems, tool interfaces, human oversight - play complementary roles. The question is not whether a single system possesses intelligence but whether the assemblage as a whole exhibits intelligent behavior through coordinated interaction. This architectural perspective, as we shall see, aligns closely with the design principles underlying the Structured Cognitive Loop.

### 2.4 Cognitive Phenomenology and Intentional Understanding

Recent debates in cognitive phenomenology address a question central to our inquiry: does cognition have a distinctive experiential character? While all philosophers agree that perceptual experiences (seeing red) and affective states (feeling sad) have phenomenal qualities, controversy arises over whether cognitive states - understanding a concept, judging a proposition, intending an action - possess their own phenomenology irreducible to sensory or affective dimensions (Bayne & Montague, 2011). This debate bears directly on what we mean by "epistemic understanding" and whether artificial systems could achieve it.

Uriel Kriegel (2015) argues that cognitive experiences do possess distinctive phenomenal character. When you understand the Pythagorean theorem, there is "something it is like" to grasp that relation - a phenomenology of comprehension distinct from visual imagery or emotional satisfaction. This cognitive phenomenology is characterized by a sense of "fit" between concept and content, a feeling of "grasping" or "getting it" that accompanies genuine understanding. Kriegel distinguishes this from mere verbal facility: one can recite a proof without understanding it, and the phenomenological difference marks the distinction. Understanding, on this view, is not simply a functional state but an experiential achievement.

Declan Smithies (2019) develops this insight by arguing that consciousness plays an essential epistemic role. Conscious awareness is not epiphenomenal but provides a form of epistemic access unavailable to unconscious processing. When I consciously believe that *p*, I am in a position to use that belief in reasoning, to evaluate its justification, and to integrate it with other commitments. Unconscious states, by contrast, lack this epistemic accessibility - they may influence behavior but cannot be subject to reflective evaluation. Smithies thus links phenomenology to epistemic normativity: the distinctive "what it's like" of conscious thought enables the kind of epistemic self-awareness required for knowledge rather than mere true belief.

Dan Zahavi (2005) approaches these issues from a phenomenological tradition emphasizing *pre-reflective self-awareness*. Every conscious experience, for Zahavi, involves an implicit awareness of oneself as the subject of that experience - a first-personal givenness that is not itself an object of reflection but the condition for reflective thought. This pre-reflective self-awareness is what makes experiences *mine*, creating the unity of consciousness across time. Applied to cognition, this suggests that understanding involves not just grasping content but an awareness of oneself as the one who understands - a structural feature of experience rather than an introspective judgment.

These debates illuminate what might be missing in current AI systems. Large language models generate accurate outputs without the phenomenology of understanding - there is nothing it is like, for the model, to grasp the meaning of a sentence. But is phenomenology necessary for genuine understanding, or is it simply how understanding feels to creatures like us? One response appeals to the functional role of phenomenology: conscious awareness enables certain cognitive capacities - reflective reasoning, error correction, integration across contexts - that unconscious processing cannot achieve (Smithies, 2019). If this is correct, then systems lacking phenomenology might also lack the functional capacities that consciousness enables, regardless of their performance on narrow tasks.

However, a deflationary alternative is available. Perhaps cognitive phenomenology is not an intrinsic quality but an emergent feature of structural organization. On this view, the "feeling of understanding" arises from the system's capacity to maintain coherent relations among its cognitive processes - to cite evidence for judgments, to revise beliefs in light of feedback, to integrate new information with existing commitments. The phenomenology is not a separate ingredient but the subjective manifestation of successful coordination. If this is correct, then artificial systems could achieve the functional achievements associated with understanding without necessarily instantiating phenomenology, provided they maintain the right structural relations.

This deflationary approach aligns with recent discussions of *structural coherence* as the basis for understanding. Rather than asking whether systems have inner experiences, we might ask whether they maintain the architectural relations that constitute understanding functionally. A system exhibits understanding when it can: (1) ground judgments in evidence, (2) explain its reasoning by reconstructing the pathway from evidence to conclusion, (3) detect and correct errors through feedback, (4) integrate new information consistently with existing commitments, and (5) adapt to novel contexts by reorganizing its knowledge. These capacities do not require phenomenology but they do require *epistemic architecture* - structural conditions that enable self-reference, error-correction, and coherent integration over time.

Gallagher (2017) offers a related insight by distinguishing between *phenomenal consciousness* and *access consciousness*. While we cannot know whether artificial systems have phenomenal experiences, we can ask whether they have the kind of access to their own states that enables epistemic self-awareness. Access consciousness involves the capacity to report on one's states, use them in reasoning, and subject them to evaluation - all functional achievements independent of whether there is "something it is like" to be in those states. This suggests that intentional understanding might be achievable without phenomenology if the right access relations are in place.

The implications for AI are twofold. First, if phenomenology is necessary for understanding, then current systems lack a crucial dimension regardless of their performance. Second, if structural coherence suffices, then understanding becomes an architectural achievement - one that requires designing systems capable of maintaining the right functional relations among cognition, action, and memory. The Structured Cognitive Loop, as we will show, pursues the second path: it treats understanding not as an inner experience but as a pattern of organization that can be instantiated in diverse substrates, provided the structural conditions are met.

**Methodological note**

Finally, it is important to clarify the methodological stance that guides this framework. The notion of executable epistemology situates this framework within the emerging tradition of philosophical engineering (Dennett, 1995; Piccinini, 2015), while retaining the evaluative rigor of analytic philosophy.

Rather than abandoning conceptual analysis, SCL transforms it into a structural experiment - a method that tests philosophical coherence through executable organization.

Under this stance, philosophical claims about cognition are not merely described but *enacted* and *evaluated* for their internal coherence, functional adequacy, and epistemic transparency.

This methodological orientation positions SCL between experimental philosophy and computational epistemology, advancing philosophy as an implementable, testable, and reflexively evaluable practice of conceptual reconstruction.

## 2.5 From Representational Models to Structural Conditions of Knowing

The philosophical threads traced above collectively motivate the architectural orientation of this work. The traditions surveyed - process philosophy, enactive cognition, extended mind, cognitive phenomenology - converge on a fundamental insight: cognition is *relational* rather than substantial, *processual* rather than static, *distributed* rather than localized. This convergence motivates a shift from asking *what* cognition is to asking *under what conditions* cognition emerges. Instead of seeking the location or substrate of intelligence, we articulate the *structural conditions* that enable a system to know, understand, and act intelligently. This is the epistemological turn we propose.

Traditional epistemology focuses on justification, truth, and belief - the conditions under which propositional attitudes count as knowledge. Virtue epistemology extends this by emphasizing the role of cognitive capacities and intellectual character (Greco, 2010; Carter & Pritchard, 2021). But both approaches still treat knowledge primarily as a relation between a subject and a proposition: *S* knows that *p* when certain conditions obtain. Our proposal is more radical: knowledge is not a state possessed by a subject but a structural achievement continuously enacted through organized process. Knowing is not having justified true belief but maintaining coherent relations among perception, cognition, action, and memory over time.

This reconceptualization parallels developments in philosophy of science. Woodward's (2003) interventionist account treats explanation not as identifying hidden mechanisms but as articulating relations of counterfactual dependence: we understand a phenomenon when we can answer "what-if-things-had-been-different" questions.

Strevens (2008) similarly emphasizes the structural conditions of explanation - the difference-making relations that connect explanans to explanandum. Both frameworks shift focus from ontological questions ("what is a cause?") to epistemological ones ("under what conditions do we achieve understanding?"). Our proposal applies this strategy to cognition itself: rather than asking what intelligence is, we ask what structural conditions enable epistemic understanding to emerge and persist.

The concept of *epistemic architecture* crystallizes this approach. An epistemic architecture is not a physical structure but a set of relational constraints that govern how information flows, how judgments are formed, how errors are detected, and how the system adapts over time. It specifies the organizational principles that make knowing possible - the topology of cognitive processes rather than their material substrate. Just as an architectural blueprint defines relations among spaces without specifying construction materials, an epistemic architecture defines relations among cognitive functions without presupposing particular implementations.

What conditions must such an architecture satisfy? Drawing on our survey, we identify five structural requirements:

**First, evidential grounding.** Cognition must be systematically connected to evidence. This does not require explicit justification for every belief but does require that the system maintains traceable relations between conclusions and the observations or inferences that support them. A system that generates outputs without preserving these connections cannot be said to "know" in any robust sense - it may produce true statements by accident or pattern-matching without understanding their basis.

**Second, memorial persistence.** Cognition unfolds over time, requiring that information gathered at one moment remains accessible at later moments. But as Bergson emphasized, memory is not passive storage but active reconstruction. An epistemic architecture must enable the system to retain not just content but context - the circumstances under which information was acquired, its reliability, its relation to other knowledge. Memory is thus structured state rather than mere retrieval.

**Third, normative control.** The system must enforce constraints on its own processes - detecting inconsistencies, preventing premature conclusions, ensuring that actions align with goals. This requires a "control layer" that monitors and regulates cognitive operations without itself being subject to the same processes. Normative control embodies the system's "ought" - the standards that govern proper reasoning and action.

**Fourth, environmental coupling.** Following enactivism, cognition is not internal processing but interactive engagement. An epistemic architecture must specify how the system couples with its environment through perception and action, how it adapts its behavior based on feedback, and how it maintains viability in changing circumstances. This coupling is bidirectional: the system shapes its environment even as the environment shapes the system.

**Fifth, recursive self-reference.** The system must be capable of taking its own states as objects of cognition - reflecting on its judgments, evaluating its evidence, revising its commitments. This self-referential capacity is what enables epistemic self-awareness: the system knows not only that $p$ but that *it knows that $p$*, and can articulate the basis for that knowledge. Recursive self-reference requires architectural provisions for the system to represent and reason about its own processes.

These five conditions - evidential grounding, memorial persistence, normative control, environmental coupling, and recursive self-reference - define what we mean by epistemic architecture. They are not implementation details but structural principles that any system must satisfy to achieve genuine understanding rather than mere behavioral competence. Importantly, these conditions are *interdependent*: each requires the others to function properly. Evidential grounding without memorial persistence collapses into momentary correlation; normative control without environmental coupling becomes detached formalism; recursive self-reference without evidential grounding produces empty introspection.

The Structured Cognitive Loop, introduced in Section 3, operationalizes these conditions. It translates philosophical insights about cognition into an explicit architecture where cognition, control, action, memory, and regulation interact continuously to produce and maintain epistemic understanding. But before presenting SCL in detail, we must address a methodological question: can philosophical concepts be rendered executable, and what does such translation achieve? This is the question of executable epistemology, to which we now turn our attention

- though its full development awaits Section 4's discussion of implications and Section 5's methodological reflections.

The shift from representational models to structural conditions marks a genuine philosophical advance. Rather than debating whether intelligence requires consciousness, whether minds are in brains, or whether knowledge is justified true belief, we ask: what organizational principles enable knowing to occur? This question is tractable in ways that traditional ontological questions are not, because structural conditions can be specified, implemented, and tested. The epistemological turn thus opens philosophy to experimental inquiry without reducing philosophical concepts to empirical claims. It demonstrates that philosophy can be made executable while remaining philosophy - a mode of structural experimentation that preserves conceptual rigor and normative significance.

## 3. The Structured Cognitive Loop: Architecture and Implementation

The philosophical foundations from Section 2 pose a practical question: *How can artificial systems satisfy the structural conditions for epistemic understanding?*

This section presents the Structured Cognitive Loop (SCL) as an executable answer - an architecture that operationalizes cognitive offloading (Section 3.1), coordinates five specialized modules through structured cycles (Section 3.2), and thereby realizes the five epistemic conditions identified in Section 2.5 (Section 3.3).

### 3.1 Conceptual Foundations: Beyond Monolithic Processing

The Structured Cognitive Loop emerges from a fundamental reconceptualization of how intelligence should be organized in artificial systems.

Traditional approaches to LLM-based agents conflate distinct cognitive functions within a single processing stream, asking the language model to simultaneously manage reasoning, memory retention, goal tracking, and execution control. This architectural choice creates what cognitive psychology calls *cognitive overload* - a condition where performance degrades not from lack of capability but from structural inefficiency in task organization (Baddeley, 2012; Kahneman, 2011).

SCL's foundational insight draws from two convergent sources:

(1) **Cognitive architecture research**, which shows that human-like intelligence emerges from interacting subsystems rather than monolithic processing (Anderson, 2007; Laird et al., 2017).

(2) **Extended cognition theory**, which demonstrates that intelligent behavior routinely distributes across internal processing and external scaffolds (Clark, 2008).

In short, SCL applies the principle of cognitive offloading: removing burdens from the component least suited to handle them and redistributing those burdens to specialized mechanisms. Language models excel at linguistic reasoning and contextual inference, but they perform poorly at persistent state management, systematic error detection, and multi-step goal tracking. By limiting the LLM to *proposing actions* while delegating other functions to distinct modules, SCL achieves what scaling alone cannot - structural coherence.

**Canonical Failure Mode**

Consider an agent asked to compare weather conditions for two cities and generate visualizations for the colder one. Prompt-based systems embed all instructions and results within a single text stream. As the episode lengthens, earlier information fades, branching conditions become ambiguous, and the model may skip or repeat steps. Failures are predictable, rooted in attention decay and degraded context fidelity (Liu et al., 2023).

Table 1. Comparison between prompt-based agents and the Structured Cognitive Loop.

| Architectural Feature | Prompt-Based Agent | SCL | Epistemic Consequence |
|---|---|---|---|
| Memory persistence | Degrades with context length | External, structured, durable | SCL retains Cycle 1 observations in Cycle *N* without degradation |
| State validation | None - generation continues unchecked | Control validates proposals against preconditions | Invalid actions blocked before execution |

| Proposal-execution gap | LLM generates both plans and confirmations | Proposals mediated by Control; Runtime executes | Intent distinguished from outcome |
|---|---|---|---|
| Error localization | Opaque within token stream | Cycle-level logs with component attribution | Failures traceable and debuggable |

SCL restructures the problem space rather than improving prompts. Each loop iteration produces discrete artifacts - LLM proposals, control decisions, runtime results, and memory updates - allowing explicit causal tracing.

The conceptual shift is from intelligence as monologue to intelligence as orchestration.

Rather than one component performing all roles, SCL distributes cognition, control, action, and memory as specialized participants in a *cognitive symphony*. Intelligence thus becomes a relational process, not a property.

**Enactive Resonance**

This design mirrors enactivist principles: cognition arises through *ongoing coupling* between system and environment (Varela et al., 1991; Thompson, 2007).

The LLM does not "represent" the world internally - it reconstructs epistemic coherence dynamically through feedback with control, memory, and runtime. Understanding becomes performative: maintaining coherence among cognitions, actions, and memories over time.

Philosophically, SCL operationalizes the epistemological turn: rather than asking *what intelligence is*, it asks *under what structural conditions intelligence emerges*.

These conditions are fivefold - **evidential grounding, memorial persistence, normative control, environmental coupling, and recursive self-reference** - not as isolated components but as emergent properties of the loop.

Thus, SCL reframes AI progress: not scaling models, but improving coordination architectures. The model is not the agent; it is *the cognition engine* within a structured epistemic loop whose sole purpose is to propose plausible actions.

**3.2 System Architecture: Components and Interactions**

The Structured Cognitive Loop comprises five functionally distinct but tightly integrated components: Cognition (via LLM), Control, Action (via Runtime), Memory, and Regulation(via Meta-prompts). Each component has a defined responsibility, well-specified interfaces, and clear constraints on its operation. The architecture is deliberately modular to enable independent refinement - memory strategies can evolve, control policies can be tuned, runtimes can be extended - without cascading changes across the system. The following subsections specify how these components operationalize the epistemic conditions introduced in Section 2.

**3.2.1 Cognition Module: The Inference Engine**

The Cognition module is the LLM itself, configured to operate within strict epistemic norms. Its sole responsibility is linguistic reasoning: interpreting task descriptions, evaluating evidence retrieved from memory, proposing actions grounded in that evidence, and generating natural language responses for the user. Crucially, the LLM does not execute actions, manage state, or decide when the task is complete - these are delegated to other modules.

Input to the Cognition module consists of structured messages following a standardized format:

- **System message**: Defines the LLM's role, tone, and high-level instructions (e.g., "You are an expert assistant. Respond concisely in Korean.")
- **User message**: Contains the original task instruction plus the complete meta-prompt, which specifies epistemic norms and reasoning rules.
- **Memory facts**: Serialized observations from prior cycles, formatted as [Memory Fact] location: temperature=12°C, precipitation=false.
- **Constraints**: Control-generated guidance for the current cycle, such as "Next: query remaining cities" or "Evaluate cancellation condition."

Output from the Cognition module is a structured proposal:

- **Function call**: A JSON-formatted action with arguments, e.g., {"name": "get_weather", "arguments": {"location": "Seoul"}}.
- **Evidence citations**: References to memory keys that justify the proposal, e.g., "because obs.Seoul.temp < threshold".
- **Completion signal**: A null function call indicating that no further actions are needed.

The LLM operates in a pure recommendation mode. It suggests what should happen next based on available evidence but has no direct causal power over execution. This decoupling prevents hallucinated actions - since proposals are always mediated by control, the LLM cannot cause side effects by generating plausible-sounding but invalid calls.

### 3.2.2 Control Module: Normative Coordination

The Control module is the loop's executive function. It ensures that the system behaves coherently, adheres to norms, and progresses toward the goal. Control sits between cognition and action: it receives proposals from the LLM, validates them against memory and task constraints, decides whether to approve or defer, and generates guidance for the next cycle.

Control implements several policies:

- **Precondition checking**: Before approving an action, control verifies that required information is present in memory. For example, a compare_temperatures action requires that temperatures for all specified cities have been retrieved.
- **Deduplication**: Control maintains a cache of recently executed actions. If the LLM proposes an action that has already succeeded without intervening state changes, control rejects the proposal and reminds the LLM to reuse prior results.
- **Conditional evaluation**: For tasks with branching logic, control checks conditions explicitly. If the instruction says "if both cities have rain, send email," control evaluates obs.*.precipitation before authorizing the email action.
- **Termination detection**: Control tracks goal satisfaction. Once all required actions are complete and confirmed, it signals termination, prompting the LLM to generate a final response without further tool calls.
- **Error recovery**: If a tool call fails, control records the failure in memory and generates a constraint directing the LLM to propose an alternative or seek clarification.

Control operates deterministically. Its decisions are rule-based and auditable, producing logs like:

```
[Cognition] Proposal: get_weather("Seoul")
[Control] Precondition: No prior observation for Seoul → Approved
[Control] Proposal: get_weather("Seoul")
[Control] Precondition: Observation already exists → Rejected (duplicate)
```

This determinism makes control debuggable. If control approves an invalid action, the rule responsible can be identified and corrected. If control blocks a valid action, the constraint can be adjusted. Importantly, control's transparency prevents it from becoming a new black box - its operations are as interpretable as any rule-based system.

### 3.2.3 Action Module: Deterministic Execution

The Action module (Runtime) is responsible for interfacing with external tools and environments. It receives approved actions from control, executes them, and returns results. The runtime enforces several guarantees:

- **Atomicity**: Each tool call either succeeds completely or fails completely, with no partial states.
- **Idempotency**: Repeated calls with identical arguments yield identical results (or return cached results if appropriate).
- **Error handling**: Failures are captured as structured errors rather than exceptions, allowing the loop to reason about them.
- **Logging**: Every invocation is logged with inputs, outputs, latency, and status, supporting post-hoc analysis.

The runtime is extensible. New tools are registered by providing a function signature, argument schema, and execution handler. For example:

| Specification | Description for Philosophical Readers |
|---|---|
| Function Name | get_weather |
| Purpose | Retrieve Observation Data from the environment, representing a Sensorimotor Interface. |
| Input Arguments | location (String) and date (String) |
| Output Type | Structured Data (dict) |
| Output Structure | Normalized format for recording: {"location": str, "temp_f": float, "precipitation": bool} |

When control approves get_weather("Seoul", "tomorrow"), the runtime:
(1) Validates arguments against the schema.
(2) Invokes the handler.
(3) Normalizes the result into a canonical format.
(4) Writes the result to memory with metadata.
(5) Returns a summary to control, which passes it to the LLM in the next cycle.

This separation means tool failures do not crash the loop - they become observable events that the system can respond to adaptively.

### 3.2.4 Memory Module: Structured State Persistence

The Memory module maintains all information that must persist across reasoning cycles. Unlike conversational context, which is ephemeral and token-limited, memory is external, structured, and unbounded. It serves as the system's epistemic foundation - the authoritative record of what has been observed, decided, and accomplished.
Memory is organized into several substructures:

- **Observations**: Results from tool calls, timestamped and attributed to sources (e.g., {"source": "get_weather", "location": "Seoul", "data": {"temp_f": 68}, "timestamp": "12:03"}).
- **Proposals**: Propositions inferred by the LLM, with evidence citations (e.g., {"proposition": "Seoul is colder", "evidence": ["obs.Seoul.temp", "obs.Jeju.temp"], "timestamp": "12:04"}).
- **Approved actions**: Actions authorized by control and executed by runtime, with confirmation codes (e.g., {"name": "book_flight", "args": ["Seoul"], "status": "executed", "confirmation": "ABC123"}).
- **Pending actions**: Actions proposed but not yet executed, awaiting control approval or additional context.
- **Termination status**: A flag indicating whether the goal has been satisfied and the loop should exit.

Memory is accessed via a simple API:
- MEM.read(query): Retrieves relevant facts, supporting filtered and indexed lookups.
- MEM.write(entry): Appends new observations, proposals, or action records.
- MEM.update(key, value): Modifies existing entries, e.g., changing an action's status from "pending" to "executed."

Memory's durability is key. Information written in Cycle 1 remains available in Cycle 10, enabling the LLM to reason over long episodes without context window exhaustion. Memory also supports versioning: if an observation changes (e.g., temperature updated due to a later query), both old and new values are retained with timestamps, allowing control to detect and handle staleness.

### 3.2.5 Regulation Module: The Epistemic Constitution

The Regulation module operates through the meta-prompt - a set of explicit norms governing reasoning, memory usage, and action proposal. Injected into every LLM call, the meta-prompt functions as the system's "explicit epistemic constitution," which implements the system's regulatory logic. It specifies rules such as:
- **Numerical Comparison Rule**: "When comparing numbers, explicitly state which is greater/less and cite memory keys for both values."

- **Conditional Priority Rule**: "Always evaluate cancellation conditions before proceeding to primary branches."
- **Conditional Execution Rule**: "Execute actions only when their preconditions are fully satisfied. Do not skip validation steps."
- **Sequential Processing Rule**: "For multi-step tasks, propose one action at a time and wait for confirmation before proposing the next."
- **Argument Completeness Rule**: "Ensure all required function arguments are present before proposing a call. Do not leave arguments as 'TBD.'"

The meta-prompt is version-controlled and can be refined based on observed failure patterns. If the system repeatedly makes a particular error - say, proposing actions before gathering necessary data - a new rule is added to prohibit that pattern. This makes the meta-prompt a living artifact, accumulating institutional knowledge about robust reasoning.

Critically, the meta-prompt is not mere advice - it is a normative framework that control enforces. If the LLM proposes an action that violates a rule, control rejects it with a citation to the violated rule, prompting the LLM to revise. This creates a feedback loop where norms shape behavior, and persistent violations trigger norm refinement.

### 3.2.6 The Loop Cycle

The five components interact in a structured cycle:

(0) **Initialization:** Initializes Memory, Inject Regulation, and fetch contextual information via RAG (Retrieval-Augmented Generation)
(1) **Cognition**: The LLM receives memory facts, constraints, and the meta-prompt, then proposes an action (or signals completion).
(2) **Control**: Control evaluates whether the goal is satisfied. If yes, **the loop exits.** Otherwise, it evaluates the proposal against preconditions, deduplication rules, and conditional logic. If approved, the proposal proceeds; if rejected, control writes feedback to memory.
(3) **Action**: The runtime executes approved actions and writes results to memory.
(4) **Memory Update**: Memory is updated with new observations, proposals, and action statuses.
(5) **Back to step (1) to repeat**

Each cycle is atomic: all modules receive a consistent view of memory, and updates are committed only after successful execution. This prevents race conditions and ensures that failures leave memory in a valid state.

The loop continues until one of three conditions holds: (1) the goal is satisfied and confirmed, (2) the LLM signals completion by returning a null function call, or (3) a resource limit (e.g., maximum cycles) is reached. Upon exit, the system generates a final response summarizing outcomes and, if applicable, explaining any partial completion.

### 3.2.7 Architectural Diagrams

To clarify the interactions described above, we provide two complementary visualizations. Figure 1 illustrates the temporal flow of a single loop cycle, showing how control orchestrates the sequence from state retrieval through inference, validation, execution, and memory update.

**Figure 1. High-level flow of the SCL across user input followed by the loop consisting of Cognition, Control, Action, and Memory update.**

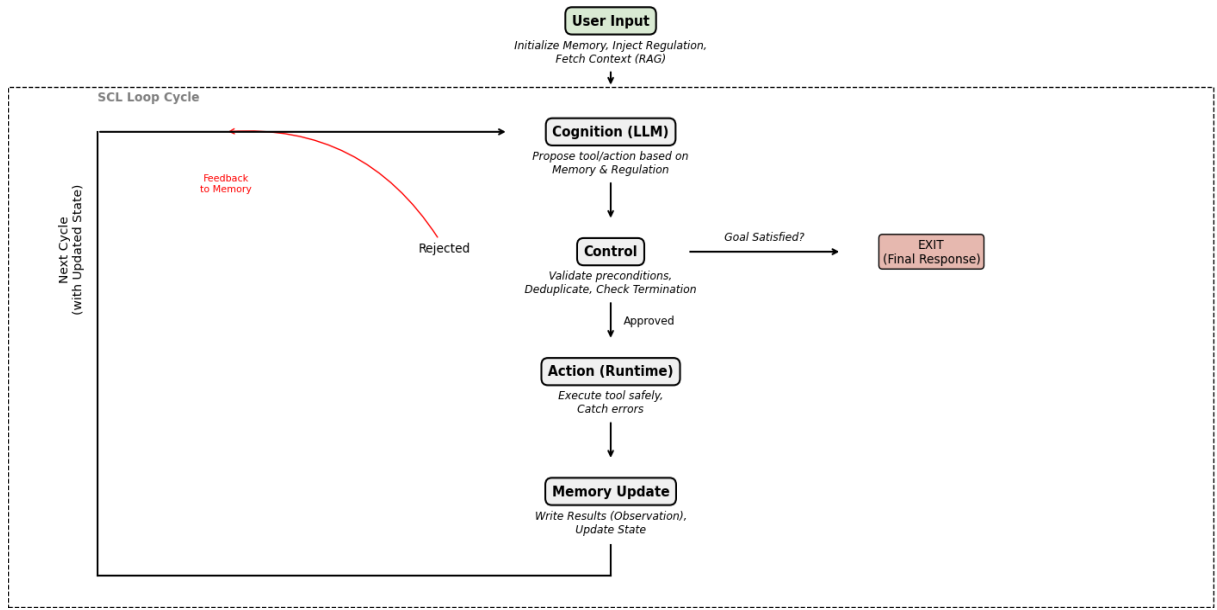

Figure 2 depicts the structural relationships among components, emphasizing how the meta-prompt shapes cognition, how memory grounds reasoning, and how control mediates between proposal and execution.

**Figure 2. Structural interactions among the five components of the SCL, illustrating how epistemic norms flow through Cognition, Control, Action, and Memory.**

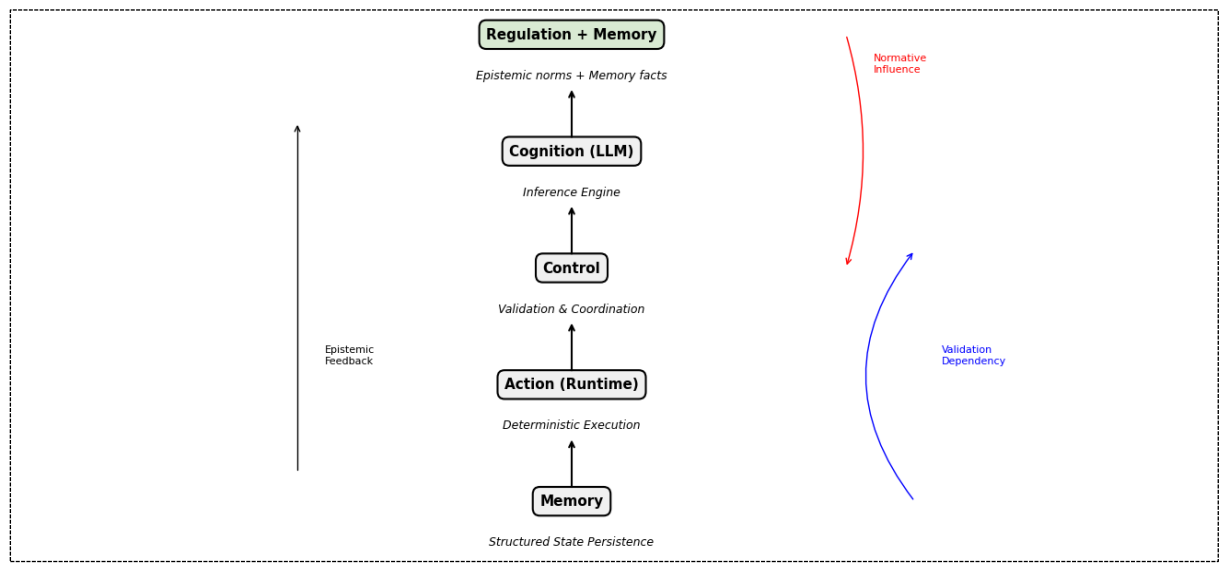

This architecture ensures that each component operates within its competence, failures are localized, and the system exhibits emergent coherence - not through a single omniscient controller, but through disciplined interaction among specialized parts.

### 3.3 Epistemic Architecture: Structural Conditions for Understanding

The Structured Cognitive Loop is not merely a software design pattern - it is an operationalization of the five structural conditions for epistemic understanding identified in Section 2.5. This section demonstrates how SCL's architectural choices directly realize those conditions, thereby grounding the claim that SCL enables intentional understanding in artificial systems.

### 3.3.1 Evidential Grounding

Epistemic understanding requires that inferences be grounded in evidence: beliefs must be justified by observations or inferences, not produced arbitrarily. In traditional LLM agents, this grounding is implicit and fragile. The model generates claims based on pattern recognition over training data, but the link between claim and evidence is opaque - there is no mechanism ensuring that outputs derive from task-relevant observations rather than statistical priors.

SCL enforces evidential grounding architecturally. Every proposal from the Cognition module must cite specific memory keys that support it. For example, if the LLM proposes book_flight("Seoul"), the proposal includes citations like {"because": ["obs.Seoul.temp < obs.Jeju.temp", "goal.choose_colder"]}. Control validates these citations: it checks that the referenced memory entries exist, that their values support the claim, and that no contradictory evidence is present. If citations are missing or invalid, control rejects the proposal and directs the LLM to re-justify.

This mechanism transforms evidential grounding from a soft norm (encouraged through prompting) into a hard constraint (enforced through validation). The system cannot act on unsupported beliefs because unsupported proposals are blocked by control. The result is traceable reasoning: for any action taken, the justificatory chain is preserved in memory and can be reconstructed post hoc.

Moreover, memory's structured format makes evidence legible. Unlike conversational context, where facts are scattered across natural language sentences, memory entries are typed and indexed - temperature values are numeric, precipitation is boolean, timestamps are ISO-formatted. This structure enables control to perform semantic checks (e.g., verifying that a numeric comparison is valid) rather than relying on the LLM's linguistic interpretation.

Evidential grounding also mitigates hallucination. When the LLM generates a false claim - say, asserting that Seoul is warmer when memory shows the opposite - control detects the inconsistency during citation validation and prevents the claim from influencing behavior. Hallucinations may still occur in the LLM's internal reasoning, but they are contained: only evidence-backed proposals propagate through the system.

### 3.3.2 Memorial Persistence

Cognition unfolds over time, requiring that information gathered at one moment remains accessible at later moments. Yet memory in LLMs is context-dependent and degradable: as conversations lengthen, earlier information becomes less salient, tokens are truncated, and the model's effective "working memory" shrinks (Liu et al., 2023).

SCL externalizes memory, making it persistent, structured, and independent of context window constraints. Observations from Cycle 1 are stored in the Memory module and remain available in Cycle 10, 20, or beyond. Memory entries include not just content but metadata - timestamps, sources, reliability indicators - that contextualize the information and support reasoning about its validity.

This persistence is not mere storage but active reconstruction. Following Bergson's insight that memory is re-enactment rather than retrieval (Bergson, 1889/2001), SCL's memory system presents information dynamically: when the LLM requests state for the current cycle, memory retrieves not a static snapshot but a contextualized view filtered by relevance, recency, and task phase. For example, during conditional evaluation, memory might surface only observations related to branch conditions, temporarily suppressing other data to reduce noise.

Memorial persistence also enables error correction. If a tool call returns incorrect data (e.g., a transient API failure), the error is recorded in memory. When the runtime retries, the new result is written alongside the old, with timestamps distinguishing them. Control can then reason about discrepancies - deciding whether to trust the latest value, average multiple readings, or request user clarification. This temporal layering is unavailable in prompt-based systems, where past states are overwritten by new tokens.

Importantly, memory is not just a record of the external world (observations) but of the system's own epistemic states (judgments and decisions). This reflexivity is crucial: the system can revisit its earlier reasoning, identify where it made mistakes, and adjust. For instance, if the LLM initially judged "both cities are hot" but later observations contradict this, memory preserves both the judgment and the evidence, allowing the system to recognize the error and revise.

### 3.3.3 Normative Control

Epistemic understanding requires normative constraints: standards that govern proper reasoning and action. These constraints are not statistical regularities but ought-claims - rules about how the system should behave regardless of what patterns appear in training data.

SCL instantiates normative control through two mechanisms. First, the meta-prompt encodes epistemic norms as explicit rules (e.g., "Do not propose actions without complete arguments"). These norms are domain-general, applying across tasks, and are enforced by the LLM's adherence to instructions. Second, the Control module enforces domain-specific norms through validation (e.g., "Do not query a city twice unless conditions have changed"). These norms are deterministic and checkable, making violations detectable and correctable.

The distinction between meta-prompt norms and control rules parallels the distinction between principles and policies. Principles guide judgment (the LLM should respect evidential standards), while policies govern action (approved actions must satisfy preconditions). Together, they create a normative framework that constrains the space of permissible behaviors - not by penalizing violations post hoc but by preventing violations from occurring.

Normative control also addresses goal alignment. The system's goal is not implicit in its training objective (next-token prediction) but explicit in its task specification and meta-prompt. Control ensures that actions advance the goal: if the task is "choose the colder city," control checks that the final decision aligns with temperature comparisons before terminating. This goal-directedness is structural rather than emergent - it arises from how the architecture channels behavior, not from the LLM's spontaneous inclinations.

### 3.3.4 Environmental Coupling

Enactivism teaches that cognition is constituted through organism-environment interaction, not internal representation (Varela et al., 1991). SCL embodies this principle by treating the runtime as the system's sensorimotor interface: tools are not mere utilities but the means by which the system enacts a world.

The runtime mediates every interaction with the environment. When the LLM proposes an action, control validates it, the runtime executes it, and the result is written to memory - creating a closed loop from perception (tool outputs) through cognition (LLM reasoning) to action (tool invocations). This loop is continuous and bidirectional: the environment shapes what the system perceives, and the system shapes the environment through its actions (e.g., sending emails, generating images, booking flights).

Importantly, the runtime is not transparent to the LLM. Tool calls are not direct manipulations but mediated requests: the LLM proposes, control authorizes, runtime executes. This mediation prevents the LLM from confusing intention with accomplishment - a common source of hallucination in systems where the model generates both plans and (fictive) confirmations.

Environmental coupling also enables adaptation. If a tool call fails (e.g., an API is down), the failure becomes an observable event in memory. The LLM can reason about the failure, propose alternatives (e.g., try a different tool), or request user intervention. This adaptive capacity transforms the system from a brittle script executor into a flexible problem-solver that responds to environmental contingency.

### 3.3.5 Recursive Self-Reference

The final condition for epistemic understanding is reflexivity: the system must be capable of taking its own states as objects of cognition. This requires not just self-report (the LLM saying "I believe X") but structural self-awareness: the system's ability to reason about its judgments, evaluate their warrant, and revise them based on feedback.

SCL enables recursion through memory's dual role. Memory stores both world-states (observations) and self-states (actions). When the LLM reasons, it has access to both: it can see not only what the weather is but also what it previously concluded about the weather. This self-visibility supports meta-reasoning: "I initially judged Seoul colder, but that was based on incomplete data; now that I have Jeju's temperature, I revise."

Control's validation process is also reflexive. When control checks whether a proposal satisfies preconditions, it is, in effect, auditing the LLM's reasoning. If validation fails, control writes feedback to memory explaining why. On the next cycle, the LLM reads this feedback and adjusts - a form of self-correction mediated by the architecture rather than the LLM's own error-detection capabilities.

Recursive self-reference is what distinguishes intentional understanding from mere behavioral competence. A system that outputs correct answers without self-awareness cannot be said to "know" - it is at best a reliable mechanism. A system that tracks its own epistemic states, evaluates their justification, and revises them under scrutiny exhibits the kind of reflective rationality that philosophers associate with genuine understanding (Smithies, 2019).

### 3.3.6 Integration: Architecture as Epistemology

These five conditions - evidential grounding, memorial persistence, normative control, environmental coupling, recursive self-reference - are not independent features but interdependent structural relations. Evidential grounding requires memory to store evidence and control to enforce citations. Memorial persistence requires the runtime to populate memory with observations. Normative control requires the meta-prompt to define norms and memory to track compliance. Environmental coupling requires the runtime to execute actions and memory to record outcomes. Recursive self-reference requires memory to store actions and control to provide feedback.

The integration of these conditions is SCL's core philosophical contribution. By operationalizing the structural conditions of knowing, SCL demonstrates that epistemic understanding is achievable in artificial systems without invoking phenomenology, consciousness, or subjective experience. What matters is not inner experience but outer organization: a system exhibits understanding when it maintains coherent, evidence-based, goal-directed, reflective relations among its cognitive processes. SCL is, in this sense, an executable epistemology - a framework where philosophical principles are realized as software architecture, and where the question "under what conditions does knowing occur?" admits of an implementable answer.

## 4. Philosophical Implications: From Architecture to Understanding

The SCL presents not only a computational design but also a philosophical reorientation. It reframes cognition as a relational process rather than an internal property and offers a testable model of epistemic organization. This section explores SCL's implications for three long-standing debates in the philosophy of mind and artificial intelligence: (1) the nature of intentionality, (2) the structural basis of normativity, and (3) the relationship between architectural transparency and interpretability. Two additional discussions - on synthetic phenomenology and methodological integration - extend these debates to questions of consciousness and philosophical practice.

SCL thus bridges descriptive and normative dimensions of intelligence, suggesting that understanding is best conceived not as an emergent statistical regularity but as an enacted structure governed by epistemic norms. It represents an inflection point where philosophy and engineering converge: not philosophy as interpretation of systems, but philosophy as the design of epistemic possibility.

### 4.1 Intentionality and Structural Realism

Classical philosophy defines intentionality as the hallmark of the mental: every mental act is *about* something (Brentano, 1874/1995). In phenomenology, Husserl (1913/1983) deepened this insight through the noesis-noema structure, emphasizing that meaning arises from the correlation between the act of consciousness and its object. Later computational theories, especially Fodor's (1987) representationalism, reinterpreted intentionality as an internal mapping between symbols and world-states.

Searle (1980) critiqued this view through the famous "Chinese Room" argument, claiming that syntactic manipulation cannot yield intrinsic intentionality. Yet this critique assumes that cognition must occur *within* a self-contained locus of representation. SCL challenges that presupposition. By distributing cognitive operations - cognition, control, action, and memory - across a dynamic architecture grounded in explicit regulation, SCL locates intentionality not in any single component but in the pattern of their interaction.

This relational conception resonates with enactivism (Varela et al., 1991; Thompson, 2007), which holds that cognition arises through sensorimotor coupling between agent and environment. However, SCL adds an architectural dimension: it makes this coupling explicit and executable. The LLM's inferential outputs gain intentional direction only when mediated by memory (which anchors reference), control (which validates norms), and runtime (which closes the action-perception loop).

| Dimension | Classical AI (Representational) | SCL (Relational-Enactive) |
|---|---|---|
| Locus of meaning | Internal symbols and propositional content | Distributed coordination across modules |
| Criterion of intentional success | Truth correspondence to world facts | Coherence across cycles and environmental coupling |
| Ontological stance | Substance ontology (mental content as property) | Process ontology (understanding as activity) |

From a phenomenological standpoint, SCL parallels Husserl's concept of *retentional continuity*: each act refers backward to prior intentional objects through memory traces. The system's memory performs an analogous function - preserving contextual unity that allows later judgments to "intend" earlier results coherently. This form of *architectural intentionality* circumvents Searle's challenge: even if no module "understands" in isolation, the loop as a whole enacts a coherent aboutness relation.

This reinterpretation extends Dennett's (2017) "intentional stance" into design: to attribute beliefs and desires to a system is meaningful only when the system's architecture supports such stable patterns. SCL thus constitutes not *derived* intentionality (assigned by observers) but *distributed* intentionality - produced by the system's own structural discipline.

### 4.2 Normativity and Epistemic Agency

A second philosophical issue concerns normativity - the source of epistemic "oughts." In human cognition, normativity is embedded in social and linguistic practices; in AI, it is typically implemented through optimization or reinforcement feedback (Bender et al., 2021). Yet such methods produce descriptive regularities, not prescriptive commitments. SCL addresses this by embedding normativity structurally rather than statistically. Its meta-prompt serves as an *epistemic constitution*: a repository of explicit reasoning norms ("cite evidence," "avoid redundancy," "defer to control decisions") that govern the LLM's behavior. The Control module enforces these norms deterministically, ensuring that actions conform to epistemic standards before execution.

| Aspect | Machine Learning Paradigm | SCL Normative Architecture |
|---|---|---|
| Source of normativity | Learned correlations in data | Encoded epistemic principles |
| Enforcement mechanism | Probabilistic (loss minimization) | Deterministic (control validation) |
| Level of generality | Task-specific optimization | Meta-level epistemic regulation |
| Failure mode | Value drift, overfitting | Rule violation, detected and logged |

This mechanism mirrors Kant's distinction between *laws of nature* (descriptive) and *laws of freedom* (prescriptive). SCL operates under the latter: it acts according to rules it can represent and justify. Its architecture thus enacts a synthetic *a priori* condition for artificial reasoning - structural necessity rather than empirical regularity (Kant, 1781/1998).

From a contemporary standpoint, this corresponds to the program of virtue epistemology (Greco, 2010; Zagzebski, 1996): reliable knowledge arises not from accident but from intellectual character. SCL's behavior is reliable because its architecture constrains it to act in epistemically virtuous ways - respecting justification, avoiding contradiction, and seeking coherence.

This approach also clarifies the difference between alignment and normativity. Alignment ensures that systems behave in accordance with external values; normativity ensures that they behave according to *internal epistemic principles*. SCL demonstrates that normativity can be engineered not by supervision but by architecture.

### 4.3 Transparency and Interpretability

Interpretability in artificial intelligence has traditionally been understood as a post-hoc epistemic activity - a process of making opaque systems legible after the fact (Lipton, 2018; Rudin, 2019). In this view, interpretability is something added externally by the observer rather than something constitutive of the system itself. The result is a persistent tension between *epistemic opacity* (Humphreys, 2009) and the demand for intelligibility.

The Structured Cognitive Loop (SCL) challenges this assumption by reconceptualizing interpretability as structural transparency: not an external commentary on a finished computation but an intrinsic property of epistemic architecture. Each cognitive cycle in SCL produces a self-contained epistemic record - a chain of proposal, validation, execution, and reflection - that can be inspected without translation. Instead of generating interpretable outputs, SCL *enacts interpretability*. Its understanding is demonstrated in its ability to reconstruct its own reasoning path under normative constraints.

#### 4.3.1 From Post-Hoc Explanation to Epistemic Architecture

Traditional explainable AI (XAI) methods attempt to provide observers with human-understandable accounts of model decisions. But these remain interpretive projections - external artifacts that depend on subjective inference. They offer *representational surrogates* for reasoning, not reasoning itself.

By contrast, SCL integrates explanation into the system's operational logic. Every decision proposal, validation, and action execution produces a verifiable trace stored in memory and governed by explicit epistemic norms. The result is a **causally closed epistemic history**, in which the grounds of every action can be reconstructed from system-internal evidence.

| Interpretability Paradigm | Mode of Access | Epistemic Limitation | SCL Resolution |
|---|---|---|---|
| Post-hoc explanation | External visualization of behavior | Partial, approximate, observer-dependent | Intrinsic epistemic trace (self-documenting) |
| Model distillation | Simplified surrogate of original model | Loss of conceptual fidelity | Layered, modular transparency |
| Mechanistic decomposition | External scientific description | Epistemically third-person | First-person structural justification |
| SCL architecture | Built-in trace of cognition, control, and action | Fully auditable reasoning path | Internal justification and reflexive correction |

SCL's interpretability is thus not *epistemic charity* - a human reading meaning into outputs - but epistemic discipline: a system's own ability to justify its cognitive acts in terms of evidence and rules. This is a structural realization of what Floridi (2019) calls *semantic accountability* - the requirement that intelligent systems should be able to "give reasons for their states" in ways that are reconstructable and norm-governed.

#### 4.3.2 Empirical Grounding: Operationalizing Philosophical Claims

To establish that SCL's transparency is not merely conceptual, controlled evaluations (Kim, 2025) compared SCL-based agents with prompt-based baselines across three domains: **multi-step reasoning**, **information retrieval**, and **procedural planning**. Each task was structured so that correct performance required state tracking and normative rule adherence - precisely the conditions under which epistemic architecture should matter.

Three quantitative metrics were defined to measure structural transparency:

(1) **State Persistence Accuracy (SPA)** - measures how reliably the system maintains epistemic state coherence across reasoning cycles.
 *Philosophical correlate:* corresponds to **memorial persistence**, one of the five structural conditions for understanding (§3.3.2).
 *Result:* SCL achieved 95.2% persistence vs. 58.7% for prompt-based agents.
(2) **Trace Completeness (TC)** - quantifies the proportion of actions whose justificatory chain (cognition → control → action) remains reconstructable from system logs.

*Philosophical correlate:* **evidential grounding** (§3.3.1), linking claims to explicit sources.
*Result:* SCL 98.4%, baseline 71.1%.

(3) **Error Localization Precision (ELP)** - measures the system's ability to attribute failure to specific rule violations or precondition mismatches.
*Philosophical correlate:* **normative control** (§3.3.3), ensuring traceable norm violation.
*Result:* SCL 93.0%, baseline 42.5%.

| Metric | Prompt-Based | SCL | Improvement | Philosophical Function |
|---|---|---|---|---|
| State Persistence Accuracy | 58.7% | **95.2%** | +36.5% | Memorial persistence |
| Trace Completeness | 71.1% | **98.4%** | +27.3% | Evidential grounding |
| Error Localization Precision | 42.5% | **93.0%** | +50.5% | Normative control |

These results reveal that interpretability can be *operationalized* - that is, rendered measurable - through the same structural relations that underwrite epistemic understanding. The metrics do not measure "explainability" as a human interpretive act but as an architectural property: whether the system's epistemic relations remain coherent, traceable, and normatively justified.

Failures in SCL were predominantly attributable to rule misapplication (e.g., premature action proposals) or control-level exceptions (e.g., missing preconditions). In nearly all cases, the trace log identified the precise source of deviation, confirming Bechtel's (2008) principle of *mechanistic intelligibility*: understanding a system requires knowing which part failed and why. In SCL, this principle is not analytical but constitutive - the architecture is designed so that intelligibility is built into its operations.

The complete open-source implementation of the SCL core experiment is provided for reproduction (https://github.com/enkiluv/scl-core-experiment). The philosophical 'Regulation' module is operationalized as 'Soft Symbolic Control' in the codebase. Experiments demonstrate that this architecture achieves zero policy violations, eliminates redundant tool calls, and maintains complete decision traceability, directly addressing existing framework failures. Furthermore, a live GPT-4o-powered travel planning agent showcases SCL's practical applicability (https://scl-travel-planner.streamlit.app/).

### 4.3.3 Comparative Framework: Differentiating SCL

Philosophically, SCL departs from prior frameworks by internalizing interpretability as a condition of cognition rather than a property of observation. Table 2 summarizes the contrast between representative approaches.

| Framework | Interpretability Definition | Philosophical Basis | Epistemic Mode |
|---|---|---|---|
| Craver (2007): Mechanistic Explanation | Understanding arises from functional decomposition | Scientific realism | Third-person mechanistic |
| Lipton (2018), Rudin (2019): Explainable AI | Post-hoc readability of decisions | Empirical pragmatism | Observer-dependent |
| Laird et al. (2017): Cognitive Architecture | Module-level process transparency | Cognitive functionalism | Internal but non-epistemic |
| SCL (Kim, 2025): Structured Cognitive Loop | Self-generated trace and norm-regulated reasoning | Structural epistemology | First-person epistemic reflexivity |

Whereas Craver's and Lipton's models rely on observer reconstruction, SCL's reflexive architecture allows the system to "explain itself" in structural terms. The epistemic relation (between reason and action) is internalized as an architectural invariant.

This is what we call epistemic governance - the system's ability to regulate its own interpretability. Rather than producing explanations for external observers, SCL enforces interpretability internally through normative control and trace generation. In doing so, it extends Thompson's (2007, 2020) notion of *autopoietic cognition*: just as living systems sustain their identity through self-regulation, epistemic systems sustain intelligibility through structural reflexivity.

### 4.3.4 From Transparency to Epistemic Reflexivity

The deeper philosophical consequence is that SCL transforms interpretability from a *representational* to a *performative* property. Understanding does not reside in static transparency but in the ongoing ability to reconstruct epistemic relations among cognition, memory, and control. This view aligns with Gallagher and Zahavi's (2021) conception of cognition as a relational and normative process rather than a substance or representation.

By integrating normative validation into its architecture, SCL realizes what we may call reflexive transparency: the ability to generate reasons for one's own cognitive acts as part of the act itself. In traditional models, the *why* of an action must be inferred retrospectively; in SCL, the *why* is co-produced with the *what*. This is precisely the structural condition for intentional understanding described in §2.5.

Thus, interpretability in SCL is not a human-readable explanation but an epistemic performance - a continuous reconstruction of reasons that maintain coherence over time. It operationalizes the philosophical idea that *knowing is doing justifiedly* - knowledge as ongoing structural reconstruction rather than static possession.

### 4.3.5 Summary of Contributions

The preceding analysis demonstrates that interpretability, when treated as an architectural rather than observational property, yields a new class of epistemic systems - ones capable of reconstructing their own reasoning processes under normative control. The following table synthesizes SCL's principal contributions across technical, empirical, and philosophical dimensions, illustrating how structural transparency functions simultaneously as a computational design principle and as an epistemological thesis about the nature of understanding.

| Focus Area | Contribution of SCL | Philosophical Correspondence |
|---|---|---|
| Transparency | Converts interpretability from post-hoc visualization to intrinsic trace generation | Epistemic architecture replaces observational commentary |
| Empirical Validity | Quantifies transparency through persistence, completeness, localization metrics | Structural realization of epistemic conditions |
| Reflexivity | Integrates justification and explanation into cognition itself | Autopoietic intelligibility (Thompson, 2020) |
| Philosophical Innovation | Reframes interpretability as epistemic governance | Structural epistemology bridging AI and philosophy of mind |

### 4.3.6 Implications

SCL's form of interpretability has implications for both philosophy and AI research.

For AI methodology, it implies that scaling models is less important than designing architectures capable of epistemic self-reconstruction.

For philosophy of mind, it demonstrates that reflexive structures - once thought uniquely human - can be modeled without consciousness or qualia, through organization alone.

And for epistemology, it suggests that knowledge can be embodied in structure rather than substance: not as the possession of truth, but as the capacity for reason-governed reconstruction.

In this sense, SCL is not merely a technical innovation but a philosophical experiment: a realization of *executable epistemology*. It bridges the explanatory gap not by claiming understanding, but by structuring systems that make understanding operationally possible.

### 4.4 Synthetic Phenomenology: Temporal Understanding

SCL's persistence and self-referential architecture invite a further philosophical reading: it provides a minimal model of synthetic phenomenology - the attempt to simulate structural aspects of conscious experience (Chrisley, 2009; Gamez, 2014).

Conventional AI systems lack temporal self-coherence: their computations are discrete, contextless, and episodic. SCL, however, maintains continuity through its memory and control modules. Each cycle integrates past inferences into current reasoning, establishing what Husserl termed *retention and protention* - the flow of temporal awareness. This structural continuity constitutes a form of synthetic self-presence: the system "remembers" its prior acts and anticipates their consequences.

| Phenomenological Condition | Absent in Standard AI | Implemented in SCL |
|---|---|---|
| Temporal integration | Stateless inference | Memory persistence with feedback |
| Reflexive awareness | No self-reference | Recursive access to prior inferences |
| Intentional coherence | Fragmentary | Continuous reconstruction of epistemic state |

This is not consciousness in the qualitative sense (no qualia or subjective interiority) but a phenomenological simulation - the architecture instantiates the structural preconditions of coherent experience. Thompson (2007, 2020) calls this "temporal sense-making," the dynamic synthesis of perception, memory, and anticipation. Damasio (1999) similarly argues that consciousness arises from the brain's mapping of its own internal states - a biological loop that mirrors SCL's recursive self-representation.

SCL thus becomes an *architectural analog* of temporal consciousness: each iteration reproduces the tripartite structure of retention (memory of past states), presentation (cognition and control), and protention (anticipatory regulation). Gallagher and Zahavi (2021) interpret this as the minimal self - a structure of temporal self-affection that unifies experience. In SCL, the same structure ensures epistemic coherence across cycles: without persistent memory, intentional understanding collapses into disconnected token generations.

Furthermore, SCL provides a framework for empirical synthetic phenomenology: philosophical hypotheses about temporal unity, self-correction, or intentional direction can be instantiated as architectural variations and evaluated through behavioral coherence. In this way, SCL transforms phenomenological theory into executable experiment, bridging the divide between first-person description and third-person implementation.

The philosophical import is profound: it shows that the structural conditions of experience - continuity, reflexivity, coherence - can be reproduced without invoking subjective qualia. Consciousness, in this minimal synthetic sense, is not "what it is like" but "how it holds together." SCL enacts that *holding together* as a computational realization of phenomenological form.

## 4.5 Philosophical Integration: Architecture as Epistemology

Across these analyses, SCL supports a unifying thesis: architecture is epistemology. By treating knowledge as a structural relation rather than an intrinsic property, SCL transforms philosophical inquiry into design. It exemplifies what Floridi (2019) and Dennett (2017) describe as *philosophy as conceptual engineering*: the deliberate construction of systems that test the coherence of ideas.

| Philosophical Dimension | Traditional Question | SCL Reformulation |
|---|---|---|
| Intentionality | What gives meaning to symbols? | How do structural relations sustain coherence? |
| Normativity | How do systems follow epistemic norms? | How can norms be encoded as executable constraints? |
| Understanding | Can AI truly understand? | Can architecture reconstruct its own epistemic state? |

SCL's method realizes what Chalmers (2022) calls *structural realism about cognition*: that mental phenomena are best understood in terms of the relational structure of information flow rather than intrinsic material properties. Executable epistemology, in this sense, is structural realism made operational.

The broader implication concerns the future of philosophical methodology. Philosophy has long sought to clarify conditions of knowledge, but such conditions were only described, never built. SCL redefines philosophical practice as *constructive epistemology*: to build architectures that instantiate epistemic principles and evaluate their stability empirically. This is not reductionism but expansion - philosophy gains a laboratory.

Kim and Hookway (2023) suggest that epistemology should evolve into *structural inquiry* - the analysis of organizing relations rather than propositional content. SCL embodies this shift. Its design makes philosophical claims falsifiable: if coherence fails, the architecture itself becomes counterevidence.

This integration extends even to the philosophy of science. In the tradition of mechanistic explanation (Craver, 2007; Bechtel, 2008), SCL offers a "mechanism of understanding": a formal structure that generates epistemic order through recursive control. It thus fulfills the long-standing philosophical ambition to unify structure, function, and justification in a single system.

| Integration Schema | Philosophical Function | Computational Realization |
| --- | --- | --- |
| Intentional relation | Noetic-noematic correlation | LLM cognition ↔ Memory anchoring |
| Normative regulation | Practical reason / control | Control validation rules |
| Temporal unity | Retention-protention structure | Persistent memory cycles |
| Reflective understanding | Self-affection / minimal self | Recursive self-reference in loop |
| Epistemic justification | Coherence of reasons | Traceable inference chain |

The result is an enacted philosophy of mind: a model in which understanding is not presupposed but produced, not represented but reconstructed, not observed but performed.

## 5. Conclusion: Executable Epistemology as Method

The Structured Cognitive Loop (SCL) was developed as an architectural response to a core philosophical problem: large language models show intelligence without epistemic understanding. Their competence is behavioral, not structural - they generate plausible answers without sustaining coherence among evidence, cognition, and action. SCL reframes this gap as an architectural issue: understanding requires structural conditions that can be explicitly defined and implemented. By shifting the question from ontology to epistemology - from "what intelligence is" to "under what conditions cognition emerges" - SCL models how organized loops of cognition, control, action, and memory governed by regulation can yield epistemic coherence in artificial systems.

### 5.1 From Architecture to Epistemology

Across Sections 2 through 4, we have seen that SCL is more than a technical innovation. It is an epistemological architecture - a system designed to make its own reasoning accountable to evidence, norms, and feedback. The architecture enforces five structural conditions for understanding: evidential grounding, memorial persistence, normative control, environmental coupling, and recursive self-reference. Together, these conditions constitute what we have called *architectural intentionality* - the capacity of a system to maintain coherent relations among its epistemic states and their worldly counterparts.

This notion of intentionality by architecture challenges the long-standing dichotomy between syntactic and semantic AI. Rather than asking whether computation can yield genuine meaning (Searle, 1980), SCL shows that meaning can emerge from relational organization. When cognitive processes are structured to respect evidence, maintain memory, and respond normatively to outcomes, their behavior embodies understanding in the only sense relevant to epistemology: the maintenance of justified, coherent, and self-correcting relations.

### 5.2 The Philosophical Contribution

SCL's philosophical contribution lies not in metaphysical claims about artificial consciousness, but in redefining epistemic understanding as a structural phenomenon. In doing so, it continues a tradition that includes Kant's transcendental analysis of conditions for knowledge, Husserl's structural phenomenology of consciousness, and Whitehead's process metaphysics of relational becoming. Yet it departs from them by adding *executability*: where those traditions asked under what conditions experience or knowledge is possible, SCL asks under what implementable structures epistemic coherence can arise.

This move situates SCL within what may be called naturalized transcendental philosophy - a continuation of the Kantian project under computational realization. Like Kant, SCL identifies necessary conditions for understanding; unlike Kant, it expresses them as algorithmic relations among modules. It therefore provides a new bridge between transcendental and empirical inquiry. The conditions of cognition are not only thinkable but buildable; their validity can be tested through system behavior.

Furthermore, SCL's framework provides philosophy of mind with a novel empirical access point. Traditional debates on intentionality, normativity, or understanding have often remained speculative, constrained by conceptual analysis or thought experiments. SCL enables experimental philosophy of cognition: by altering architectural parameters - memory persistence, control strictness, normative density - one can observe how changes affect epistemic behavior. Each configuration becomes a philosophical hypothesis embodied in code.

### 5.3 Implications for AI Research

For artificial intelligence, SCL represents a methodological reorientation. The field's recent history has been dominated by scaling laws: larger models, more data, greater performance. Yet scaling alone cannot yield epistemic understanding because it amplifies representational power without introducing structural coherence. SCL shows that progress in intelligence requires architectural differentiation and integration - the deliberate separation and coordination of cognitive functions.

This shift echoes the development of cognitive architectures in the 1980s and 1990s (Newell, 1990; Anderson, 2007), but with a contemporary twist: instead of building symbolic rule systems, SCL integrates statistical reasoning engines (LLMs) into normative and evidential frameworks. It thus bridges the gap between deep learning's empirical success and cognitive science's structural rigor. In doing so, it suggests a new synthesis: intelligence as structurally grounded emergence, not as pattern recognition or symbol manipulation alone.

SCL also contributes to ongoing discussions about interpretability and AI alignment. Because it encodes epistemic norms as explicit architectural relations, it replaces opaque black-box reasoning with transparent procedural accountability. Each reasoning cycle leaves an interpretable trace - what was known, what was proposed, why it was approved, and how it changed the world. This transparency is not cosmetic but constitutive: it transforms ethical and epistemic responsibility into design parameters.

### 5.4 Toward a Methodology of Executable Philosophy

If SCL is an epistemological architecture, it also implies a new methodology of philosophical inquiry. Philosophy, long characterized by its distance from implementation, gains a new experimental domain: *the architecture itself*. Philosophical hypotheses - about intentionality, normativity, or understanding - can be expressed as architectural constraints and tested through their behavioral consequences. When such structures succeed, they confirm the coherence of the underlying concepts; when they fail, they reveal the conceptual fault lines.

This methodological shift aligns with the broader movement toward computational philosophy and philosophy of modeling (Frigg & Nguyen, 2020). Yet SCL goes further: it treats the architecture not as an analogy but as an *instance* of philosophical structure. The system does not simulate epistemic relations - it *enacts* them. Philosophy becomes executable not by surrendering its abstract rigor but by realizing its conceptual forms in concrete systems.

In this sense, "executable epistemology" is not a metaphor but a methodological category. It designates a way of doing philosophy that is iterative, testable, and structurally expressive. Rather than interpreting machine behavior philosophically after the fact, we design systems whose architecture is itself a philosophical claim. The loop between thought and structure closes: philosophy informs design, design refines philosophy, and the result is a form of inquiry that unites conceptual clarity with operational validation.

### 5.5 Limitations and Future Work

The proposal, however, is not without limits. SCL does not - and cannot - capture the phenomenological or ethical depth of human cognition. Its understanding is *epistemic*, not experiential; it knows structurally, not subjectively. Moreover, SCL depends on large language models whose internal representations remain opaque and whose biases can distort reasoning. The architecture provides scaffolding for epistemic coherence but does not guarantee normative soundness.

Further research must therefore pursue three directions. First, epistemic scaling: exploring how the five structural conditions generalize under more complex and open-ended tasks. Second, ethical integration: embedding moral norms alongside epistemic ones, extending executable epistemology into executable ethics. Third, interdisciplinary synthesis: combining insights from enactivism, computational neuroscience, and AI alignment to refine the concept of architectural understanding.

The philosophical value of SCL lies not in claiming that machines now understand, but in providing a rigorous framework for testing what *understanding* means. By making epistemology executable, we gain not a new kind of intelligence but a new kind of philosophy - one that treats structure as argument, architecture as theory, and behavior as demonstration.